\documentclass[10pt,twocolumn,letterpaper]{article}

\usepackage[pagenumbers]{cvpr} 

\usepackage[dvipsnames]{xcolor}

\newcommand{\red}[1]{{\color{red}#1}}

\usepackage{graphicx}
\usepackage{float}
\usepackage{times}
\usepackage{epsfig}
\usepackage{amsmath}
\usepackage{amssymb}
\usepackage{booktabs}
\usepackage{placeins}
\usepackage{stfloats}
\usepackage[accsupp]{axessibility}  

\usepackage{commath}
\usepackage{caption}
\usepackage{subcaption}
\usepackage{verbatim}
\usepackage{gensymb}
\usepackage{rotating}
\usepackage{marvosym}
\usepackage{multirow}
\usepackage{bm}
\usepackage{enumitem}
\usepackage{makecell}

\definecolor{cvprblue}{rgb}{0.21,0.49,0.74}
\usepackage[pagebackref,breaklinks,colorlinks,citecolor=cvprblue]{hyperref}

\def\paperID{13482} 
\def\confName{CVPR}
\def\confYear{2025}

\title{Instruct-4DGS: Efficient Dynamic Scene Editing via 4D Gaussian-based Static-Dynamic Separation}

\author{Joohyun Kwon\thanks{Equal contribution.}\\
DGIST, South Korea\\
{\tt\small rnjswngus00@dgist.ac.kr}
\and
Hanbyel Cho\footnotemark[1]\\
KAIST, South Korea\\
{\tt\small tlrl4658@gmail.com}
\and
Junmo Kim\\
KAIST, South Korea\\
{\tt\small junmo.kim@kaist.ac.kr}
}

\begin{document}

\maketitle

\begin{abstract}
Recent 4D dynamic scene editing methods require editing thousands of 2D images used for dynamic scene synthesis and updating the entire scene with additional training loops, resulting in several hours of processing to edit a single dynamic scene. Therefore, these methods are not scalable with respect to the temporal dimension of the dynamic scene (i.e., the number of timesteps). In this work, we propose \textbf{Instruct-4DGS}, an efficient dynamic scene editing method that is more scalable in terms of temporal dimension. To achieve computational efficiency, we leverage a 4D Gaussian representation that models a 4D dynamic scene by combining static 3D Gaussians with a Hexplane-based deformation field, which captures dynamic information. We then perform editing solely on the static 3D Gaussians, which is the minimal but sufficient component required for visual editing. To resolve the misalignment between the edited 3D Gaussians and the deformation field, which may arise from the editing process, we introduce a refinement stage using a score distillation mechanism. Extensive editing results demonstrate that Instruct-4DGS is efficient, reducing editing time by more than half compared to existing methods while achieving high-quality edits that better follow user instructions. Code and results: \url{https://hanbyelcho.info/instruct-4dgs/}
\end{abstract}    
\vspace{-2mm}
\section{Introduction}
\label{sec:intro}
\vspace{-1mm}

\begin{figure}[!t]
\centering
    \includegraphics[width=\columnwidth]{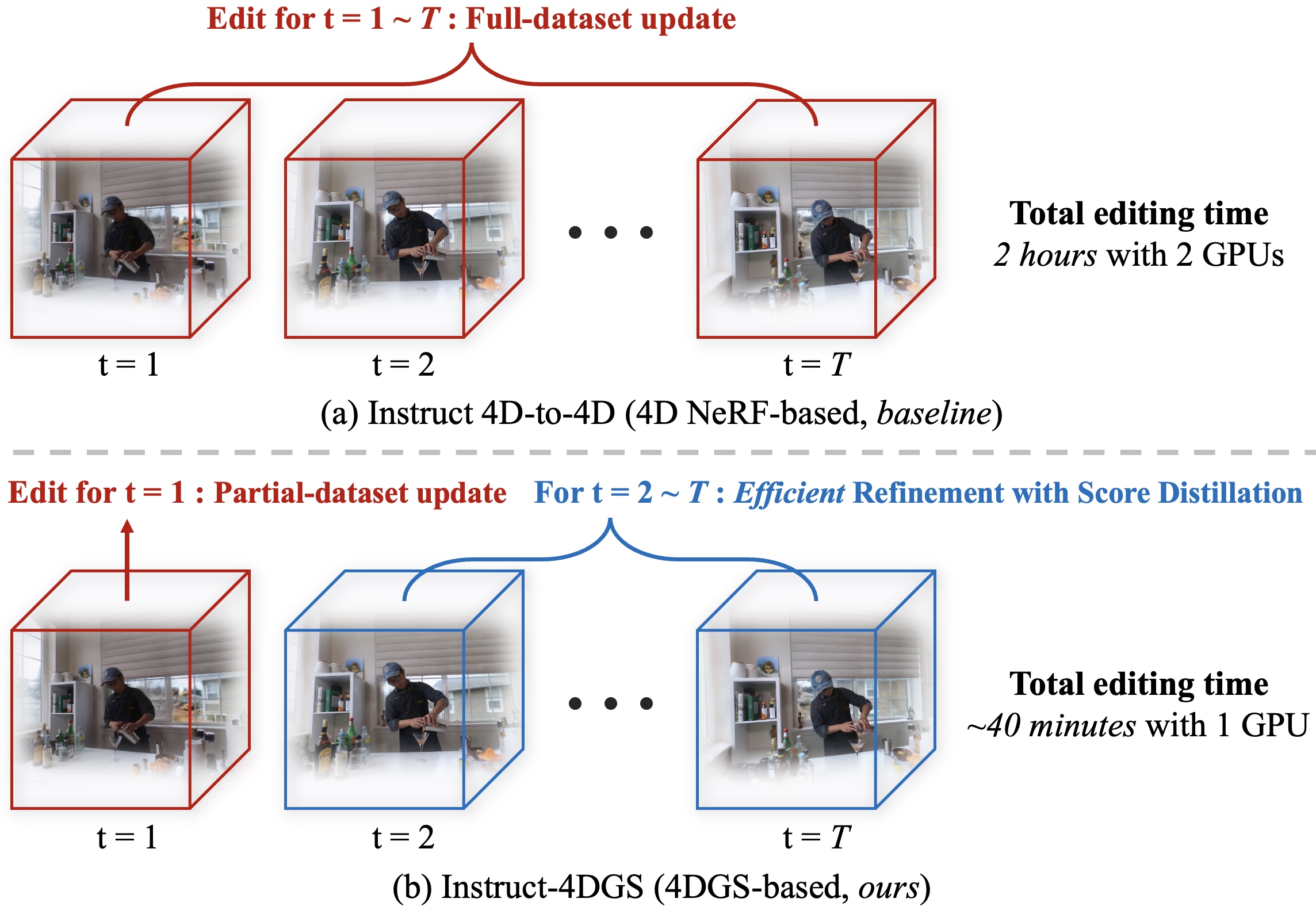}
    \vspace{-6mm}
    \caption{\textbf{Illustration of dynamic scene editing processes for baseline and our method}: (a) The existing method requires updating the 2D images for all timesteps. (b) In contrast, our method updates only the first timestep's dataset images, edits canonical 3D Gaussians, and efficiently completes dynamic scene editing through score-based temporal refinement. For a multi-camera dataset with $T=300$, our method reduces editing time by more than half compared to the baseline, using only a single GPU.}
    \label{fig:teaser}
\vspace{-4mm}
\end{figure}

Diffusion-based generative models~\cite{ref_26_ddpm, ref_2_ldm, ref_66_scalable, ref_67_plug, ref_68_controlnet, ref_69_dreambooth, ref_70_hierarchical} have recently achieved remarkable progress in the 2D image domain and are increasingly being integrated into practical applications. As the demand for generative tasks extends beyond 2D, the editing of 3D and 4D dynamic scenes has emerged as a significant area of research. In particular, user-instruction-guided editing is gaining traction as an intuitive and user-friendly approach.

In this context, InstructPix2Pix (IP2P)~\cite{ref_1_ip2p} has gained recognition by proposing a novel method for editing 2D images based on user instructions. Building on IP2P’s capabilities, research on instruction-guided 3D scene editing, particularly with NeRF~\cite{ref_4_nerf} and 3D Gaussian Splatting (3DGS)~\cite{ref_8_gs}, has become increasingly active. However, 4D dynamic scene editing remains relatively underexplored. One of the few existing methods, Instruct 4D-to-4D~\cite{ref_9_i4d24d} requires iterative dataset updates for \emph{``thousands of 2D images''} used in the dynamic scene synthesis, as shown in Fig.~\ref{fig:teaser} \red{(a)}, along with additional training loops to update the entire dynamic scene, resulting in several hours of processing to edit a single dynamic scene. Regardless of how efficiently the dataset is updated, such an approach fails to scale with the temporal dimension of dynamic scenes, making it impractical for real-world applications.

In this work, we propose \textbf{Instruct-4DGS}, an efficient 4D dynamic scene editing method that is more scalable with respect to the temporal dimension. To maximize computational efficiency, we focus on three key aspects: (1) Since 4D dynamic scenes require frequent rendering during the editing process, we employ 4D Gaussian Splatting (4DGS)~\cite{ref_10_4dgs} as our scene representation, enabling fast and efficient rendering. (2) Our objective is to edit the appearance of the scene while preserving its motion. To achieve this, we leverage the inherent separability of 4DGS into static and dynamic components—specifically, canonical 3D Gaussians (\emph{static}) and a Hexplane~\cite{ref_15_kplanes, ref_16_hexplane}-based deformation field (\emph{dynamic})—allowing us to improve efficiency by editing only the static component. (3) To ensure better alignment between the edited static 3D Gaussians and the original deformation field, we perform temporal refinement using a score distillation mechanism~\cite{ref_18_dreamfusion}.

Specifically, the Hexplane-based 4DGS offers notable advantages in both editing quality and rendering efficiency compared to the 4D NeRF~\cite{ref_22_nerfplayer} used in Instruct 4D-to-4D. By employing 3D Gaussians to represent the static canonical scene, we ensure high-quality, real-time rendering during the editing process. Additionally, Hexplane, which utilizes a spatio-temporal encoding structure based on planar factorization, is highly compact, further contributing to real-time rendering performance.

In addition to rendering efficiency, we aim to achieve computational efficiency in dynamic scene editing by focusing solely on the static component. Since our goal is to edit the scene’s appearance while preserving its motion, we modify only the static 3D Gaussians, which are the minimal yet sufficient elements for appearance editing. As shown in Fig.~\ref{fig:teaser} \red{(b)}, this approach allows us to edit the entire dynamic scene without updating every 2D images, even for scenes with extended timesteps. Specifically, we edit only a subset of 2D multiview images from the initial timestep using IP2P and then apply simple modifications to the static 3D Gaussians using L1 RGB loss.

While editing only the static 3D Gaussians is simple and efficient, it introduces motion artifacts in later timesteps. Specifically, modifying the static 3D Gaussians causes slight shifts in the positions of Gaussian primitives, leading to misalignment between the static canonical scene and the original deformation field. Additionally, only the Spherical Harmonics (SH) colors of 3D Gaussians visible in the first timestep are updated. As a result, when Gaussian primitives rotate through the deformation field in subsequent timesteps, previously unmodified SH values become exposed, introducing visual artifacts. In summary, the dynamic scene tends to overfit to the first timestep, leading to artifacts across other timesteps.

To address this temporal misalignment, we propose a refinement stage that adjusts the edited static 3D Gaussians to better align with the original deformation fields. Specifically, we utilize the score distillation mechanism proposed in DreamFusion~\cite{ref_18_dreamfusion} to transfer IP2P's editing guidance into 3D and even 4D spaces. We apply a score-based refinement stage to eliminate artifacts in the \textit{pseudo-edited dynamic scene}, where the edited static 3D Gaussians are misaligned with the deformation field. Additionally, inspired by MVDream~\cite{ref_20_mvdream} and Tune-a-Video~\cite{ref_21_tuneavideo}, we replace IP2P's self-attention module with a cross-attention module. This modified IP2P, Coherent-IP2P prevents the accumulation of non-uniform editing guidance during score distillation, which would otherwise result in blurry outputs.

Our evaluation demonstrates a significant reduction in editing turnaround time while improving visual quality. Furthermore, our method can effectively perform dynamic scene editing across various user instructions. Our main contributions are summarized as follows:
    \begin{itemize}
        \vspace{1mm}
        \item
        We propose Instruct-4DGS, the first efficient dynamic scene editing framework based on 4D Gaussian Splatting.
        \item
        \vspace{0.5mm}
        We achieve efficient dynamic scene editing by modifying only static 3D Guassians, the minimal but sufficient component for visual editing.
        \item
        \vspace{0.5mm}
        We propose a refinement method using score distillation with Coherent-IP2P, which removes motion artifacts while maintaining computational efficiency.
        \item
        \vspace{0.5mm}
        Our method reduces editing time by more than half while achieving higher visual quality.
        \vspace{0.5mm}
    \end{itemize}
\vspace{4mm}
\vspace{-4mm}
\section{Related Work}
\label{sec:formatting}

\begin{figure*}[!t]
\centering
    \includegraphics[width=\linewidth]{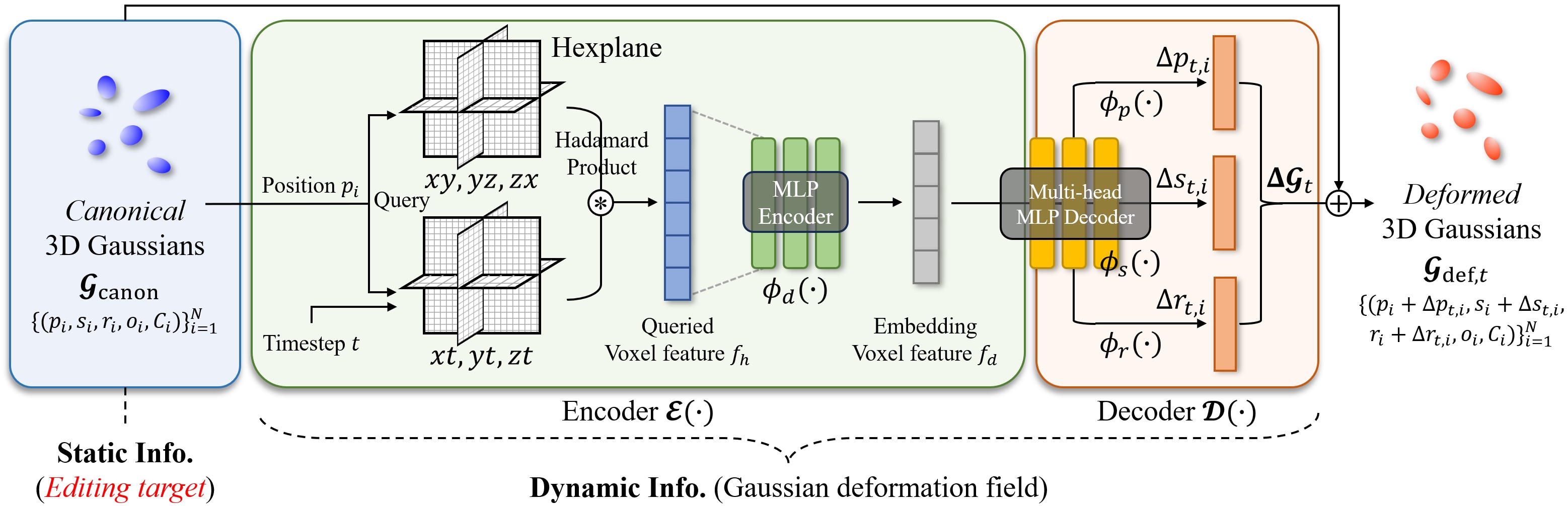}
    \vspace{-5mm}
    \caption{\textbf{Overview of 4D Gaussian Splatting}: 4DGS represents dynamic scenes by separating static (canonical 3D Gaussians $\mathcal{G}_\text{canon}$) and dynamic components (Gaussian deformation field $\left\{\mathcal{E}(\cdot), \mathcal{D}(\cdot)\right\}$). Given a Gaussian primitive's position $p$ and timestep $t$, a spatio-temporal embedding voxel feature $f_d$ is queried from the Hexplane. This feature is then processed by a multi-head MLP decoder $\{\phi_p(\cdot), \phi_s(\cdot), \phi_r(\cdot)\}$ to generate per-Gaussian deformation parameters $\Delta p_{t,i}$, $\Delta s_{t,i}$, and $\Delta r_{t,i}$. By adding these parameters to the canonical Gaussians $\mathcal{G}_\text{canon}$, we obtain the deformed Gaussians $\mathcal{G}_{\text{def},t}$. Finally, by repeatedly generating and rendering $\mathcal{G}_{\text{def},t}$ across timesteps, the dynamic scene video is obtained.}
    \label{fig:4dgs}
    \vspace{-2mm}
\end{figure*}

\subsection{4D Dynamic Scene Representation}
Recent advancements in computer vision and graphics have fueled interest in 4D dynamic scene representation, which models both spatial and temporal information. As high-quality 4D content capture continues to improve, multi-view 4D data has become increasingly available, highlighting the need for efficient representations to mitigate the high computational costs of 4D modeling. Many approaches~\cite{ref_22_nerfplayer, ref_27_tensor4d, ref_45_dnerf, ref_46_neuralradianceflow, ref_47_devrf, ref_48_nerfies, ref_49_nonrigid} have reduced the complexity of dynamic scene representation by handling the temporal dimension separately, leading to the decoupling of the canonical 3D representation and the deformation field. Specifically, K-plane and Hexplane~\cite{ref_15_kplanes,ref_16_hexplane} construct spatio-temporal encoding structures within the deformation field using multi-scale parameter grids through planar factorization. Other methods~\cite{ref_11_spacetimegs,ref_12_iclr24,ref_13_gaufre, ref_14_4drotergs} have enhanced the overall performance of dynamic scenes by employing 3D Gaussian Splatting~\cite{ref_8_gs} as the canonical 3D representation, which has recently gained attention for its real-time rendering capabilities and high visual quality. Notably, 4D Gaussian Splatting~\cite{ref_10_4dgs} combines 3DGS with the Hexplane deformation field to achieve real-time rendering speeds while more accurately modeling dynamic scenes. Given its excellent performance, 4DGS holds great potential for dynamic scene generation~\cite{ref_17_dreamgaussian4d, ref_19_aligngs, ref_25_4dfy}, editing~\cite{ref_9_i4d24d, ref_33_control4d}, and tracking~\cite{ref_50_dynamic3dgs, ref_51_motionaware}. In this paper, we employ 4DGS to maximize the efficiency of the dynamic scene editing process.

\subsection{Instruction-Guided Scene Editing}
User instructions provide one of the most intuitive and user-friendly approaches to scene editing. InstructPix2Pix~\cite{ref_1_ip2p} introduced instruction-guided editing by fine-tuning the Stable Diffusion~\cite{ref_2_ldm} model on a dataset of \textit{source image}–\textit{instruction}–\textit{target image} triplets. Recent studies~\cite{ref_3_in2n, ref_5_gaussianeditor, ref_6_dge, ref_7_tiger, ref_28_i3dto3d, ref_32_vicanerf} have extended IP2P's capabilities to 3D scenes by developing methods to ensure spatial consistency in editing guidance, thereby making significant progress in instruction-guided 3D scene editing despite the limited availability of 3D datasets. Among these approaches, one of the key trends is the iterative dataset update method, where all 2D images used for 3D scene synthesis are edited, followed by re-training the 3D scene. Recently, Instruct 4D-to-4D~\cite{ref_9_i4d24d} extended this iterative dataset update approach to 4D space, presenting the first instruction-guided 4D editing method. By employing flow-based~\cite{ref_53_raft} and depth-based warping to ensure spatio-temporal consistency during dataset updates, they achieved notable results. However, editing all images for 4D dynamic scenes remains extremely time-consuming, highlighting the need for a more efficient approach that can effectively leverage diffusion priors for dynamic scene editing. Therefore, we propose an efficient dynamic scene editing method that significantly reduces total editing time.

\subsection{Score Distillation Sampling}
 The Score Distillation Sampling (SDS) mechanism was introduced in DreamFusion~\cite{ref_18_dreamfusion} for text-to-3D scene generation, enabling the transfer of pre-trained 2D diffusion model priors~\cite{ref_1_ip2p, ref_2_ldm, ref_26_ddpm} to other data domains. When SDS is used with diffusion networks incorporating specific hypotheses—such as multiview diffusion models~\cite{ref_20_mvdream, ref_52_imagedream} or video diffusion models~\cite{ref_21_tuneavideo, ref_54_vdm, ref_55_lumiere, ref_56_makeavideo, ref_57_align, ref_58_sora}—it produces guidance that aligns with those hypotheses. Many studies~\cite{ref_17_dreamgaussian4d, ref_18_dreamfusion, ref_19_aligngs, ref_25_4dfy, ref_59_dreamgs, ref_60_dreamer, ref_61_magic3d} have leveraged this property to develop SDS-based 3D/4D generation methods. Meanwhile, other approaches~\cite{ref_7_tiger, ref_28_i3dto3d, ref_62_prolific, ref_63_dreameditor, ref_64_focal, ref_65_progressive} have explored SDS for editing tasks, adapting it to improve spatial and temporal consistency during the editing process. Furthermore, some works~\cite{ref_34_deltadenoising,ref_35_posterior,ref_36_collaborative} have enhanced editing performance by modifying the score loss function to better suit editing-specific objectives.
\vspace{4mm}
\vspace{-4mm}
\section{Preliminary}
\vspace{-1mm}
For efficient dynamic scene editing, we leverage 4D Gaussian Splatting (4DGS)~\cite{ref_10_4dgs} which represents scenes by separating static and dynamic information. In this section, we briefly review 4DGS, highlighting its Hexplane~\cite{ref_16_hexplane, ref_15_kplanes}-based Gaussian deformation field, and introduce our proposed method in Sec.~\ref{sec::method}.

\vspace{-4mm}
\paragraph{4D Gaussian Splatting.}
4DGS consists of a canonical 3D Gaussians~\cite{ref_8_gs} $\mathcal{G}_{\text{canon}}$ that represents static information and a Gaussian deformation field that represents dynamic information and produces each Gaussian’s deformation $\Delta  \mathcal{G}_t$ (where $t$ is a normalized value from 0 to 1, denoting the timestep within the dynamic scene), as illustrated in Fig.~\ref{fig:4dgs}. The 3D Gaussians $\mathcal{G}_{\text{canon}}$ representing the undeformed static canonical 3D scene consist of $N$ Gaussian primitives (in our case, $N=100k$–$200k$), denoted as $\mathcal{G}_{\text{canon}} = \left\{(p_i, s_i, r_i, o_i, C_i)\right\}_{i=1}^N$, where each primitive is defined by a position $p \in \mathbb{R}^3$ , a scaling vector $s\in \mathbb{R}^3$, a rotation quaternion $r\in \mathbb{R}^4$, an opacity $o\in \mathbb{R}$, and a spherical harmonics color $C\in \mathbb{R}^k$, with $k$ determined by the SH degree. The Gaussian deformation field $\left\{\mathcal{E}(\mathcal{G}_{\text{canon}}, t), \mathcal{D}\right\}$ consists of an encoder part $\mathcal{E}(\mathcal{G}_{\text{canon}}, t)$, which outputs an embedding voxel feature $f_d$ based on spatio-temporal input coordinates $p$ and $t$, and a decoder part $\mathcal{D}$, which decodes the voxel feature into each Gaussian’s deformation $\Delta \mathcal{G}_t$. Note that, to ensure Gaussian deformations resemble real-world physical motion, 4DGS computes deformation values only for Gaussian position $p$, scale $s$, and rotation $r$. Therefore, $\Delta \mathcal{G}_t$ can be expressed as $\left\{ \Delta p_{t,i}, \Delta s_{t,i}, \Delta r_{t,i} \right\}_{i=1}^N$. By repeatedly adding the outputs of the deformation field $\Delta \mathcal{G}_t = \mathcal{D}(\mathcal{E}(\mathcal{G}_{\text{canon}},t))$ to the canonical 3D Gaussians $\mathcal{G}_{\text{canon}}$ at each timestep $t$, we can render an image $\hat{I}_{M, t}$ from deformed 3D Gaussians $\mathcal{G}_{\text{def},t} = \mathcal{G}_{\text{canon}} + \Delta \mathcal{G}_t$ as: $\hat{I}_{M, t} = S(M, \mathcal{G}_{\text{def},t})$, where $M$ denotes the camera matrix, and $S$ represents the rendering (differential splatting) process of the 3DGS. 

\vspace{-4mm}
\paragraph{Encoder for Gaussian Deformation Field.}
4DGS incorporates Hexplane~\cite{ref_15_kplanes,ref_16_hexplane}, as a core component in the structure of the encoder $\mathcal{E}(\mathcal{G}_{\text{canon}}, t)$ within its Gaussian deformation field. The Hexplane is a spatio-temporal structure encoder and can be viewed as a generalization of Triplane ~\cite{ref_24_eg3d}, which was originally designed to embed spatial information in 3D space.

Hexplane-based encoder $\mathcal{E}(\mathcal{G}_{\text{canon}}, t)$ can be parametrized by six multi-resolution voxel grids $R_l$ across the four dimensions $(x, y, z, t)$ and simple MLP encoder $\phi_d$ as $
 \mathcal{E}(\mathcal{G}_{\text{canon}},t)=\{R_l(i,j), \phi_d | (i,j) \in \{(x,y),(y,z),(x,z),(x,t),(y,t),(z,t)\}, l \in \{1,2\}\}$, where $l$ represents the multi-resolution level (the multi-resolution technique is relevant to Instant-NGP ~\cite{ref_23_instantNGP}, enabling fast optimization and rendering).

The spatio-temporal embedding voxel feature $f_d$ is obtained from the Hexplane as $f_d = \phi_d(f_h)$, where $f_h = \bigcup_l \prod \text{interp}(R_l(i,j))$, and $(i,j) \in \{(x,y),(x,z),(y,z),(x,t),(y,t),(z,t)\}$. In 4DGS, the $x, y$, and $z$ coordinates of the Gaussian position $p$ and timestep $t$ is used to query voxel features across six planes. The six voxel features obtained from each plane through bilinear interpolation are then combined via the Hadamard product (channel-wise product). This queried voxel feature $f_h$ is subsequently passed through $\phi_d$ to yield the final embedding voxel feature $f_d$ as shown in Fig.~\ref{fig:4dgs}.

\begin{figure*}[h]
\centering
    \includegraphics[width=\linewidth]{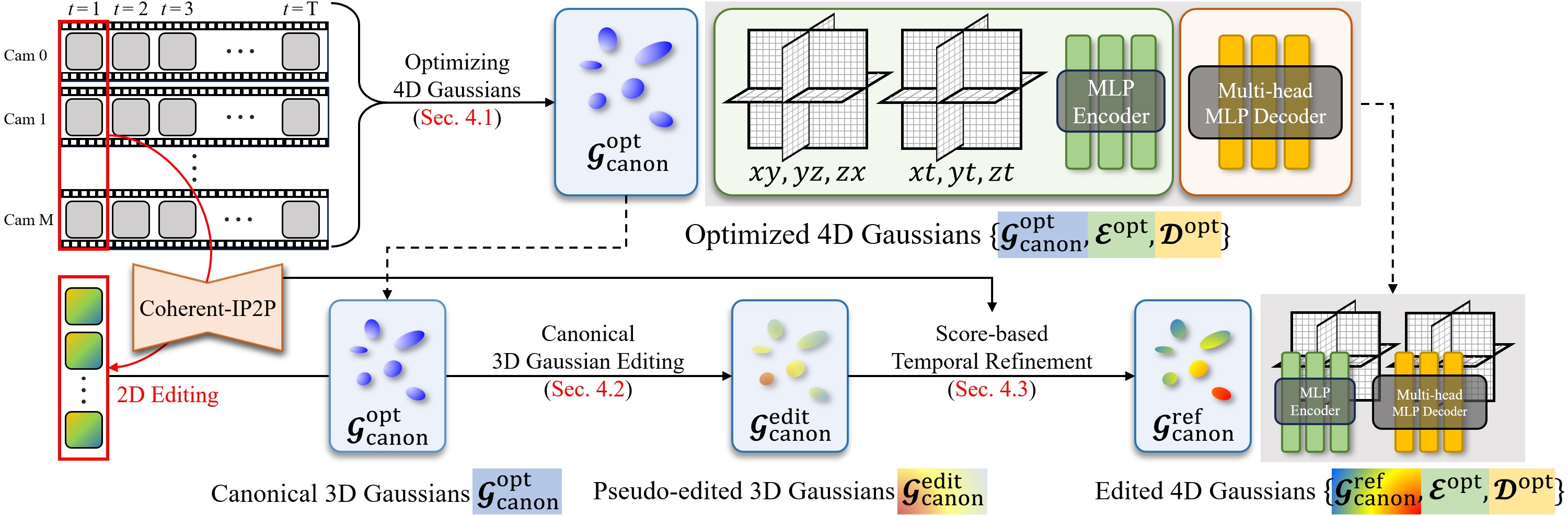}
    \caption{\textbf{Overall pipeline of our proposed dynamic scene editing method (Instruct-4DGS)}: To obtain the target dynamic scene for editing, we first optimize the 4D Gaussians using a multi-camera captured video dataset (Sec.~\ref{subsec::4.1}). We then perform 3D Gaussian editing on the static canonical 3D Gaussians by editing only the multiview images corresponding to the first timestep (Sec.~\ref{subsec::4.2}). We apply score-based temporal refinement to mitigate motion artifacts without additional image editing (Sec.~\ref{subsec::4.3}).}
    \label{fig:method}
    \vspace{-1mm}
\end{figure*}

\vspace{-4mm}
\paragraph{Decoder for Gaussian Deformation Field.}
 The spatio-temporal voxel feature $f_d$ passes through a multi-head simple MLP decoder $\mathcal{D}=\{\phi_p, \phi_s, \phi_r\}$, which decodes it into the deformation values of the Gaussian feature $p$, $s$ and $r$ as $\Delta p=\phi_p(f_d), \Delta s=\phi_s(f_d)$, and $\Delta r=\phi_r(f_d)$.
 
 Since the deformation field $\left\{\mathcal{E}(\mathcal{G}_{\text{canon}}, t), \mathcal{D}\right\}$ is designed with a compact Hexplane and a simple MLP, 4DGS achieves real-time rendering speed. This provides a significant advantage for editing tasks, where rendering is performed frequently. To leverage this efficiency and the static-dynamic separability, we applied 4DGS to our primary representation for 4D dynamic scenes.

\vspace{2mm}
\section{Method}
\label{sec::method}
In this section, we present \textbf{Instruct-4DGS}, our proposed method for efficient dynamic scene editing, as illustrated in Fig.~\ref{fig:method}. (Sec.~\ref{subsec::4.1}) We first train dynamic scenes for editing targets by optimizing the 4D Gaussian Splatting (4DGS) ~\cite{ref_10_4dgs}. (Sec.~\ref{subsec::4.2}) Motivated by the static-dynamic separability of Hexplane~\cite{ref_15_kplanes, ref_16_hexplane}-based 4DGS, we initially focus on editing the static canonical 3D Gaussians ~\cite{ref_8_gs} to efficiently edit the dynamic scene. (Sec.~\ref{subsec::4.3}) To mitigate overfitting issues that may arise during the 3D Gaussian editing process and refine the motion artifacts in the dynamic scene, we introduce a temporal refinement stage using score distillation.

\subsection{Optimizing 4D Gaussians for Target Scenes}
\label{subsec::4.1}
Our dynamic scene editing requires a 4D Gaussian representation of the target dynamic scene. To obtain this, we optimize the 4D Gaussians and use it for editing. Specifically, we use dynamic scene datasets~\cite{ref_37_neural3dvideo,ref_71_technicolor} composed of multi-camera captured videos, which can be represented as a set of images $\{I_{M,t}\}$ in which $M$ denotes the camera matrix and $t$ denotes the timestep within the videos. We synthesize images $\hat{I}_{M,t}$ by rendering the randomly initialized dynamic scene $\left\{\mathcal{G}_{\text{canon}}^\text{init},\mathcal{E}^\text{init}(\mathcal{G}_{\text{canon}}, t), \mathcal{D}^\text{init}\right\}$. Then we calculate the RGB L1 loss against the corresponding dataset image $I_{M,t}$, training the dynamic scene through this process. We also apply a grid-based total variational loss~\cite{ref_15_kplanes,ref_16_hexplane,ref_41_directvoxel,ref_42_fastdynamic}, $\mathcal{L}_\text{TV}$ to enforce smoothness in the deformation field output along the timesteps. Note that, as our editing method is highly dependent on the quality of the target dynamic scene, incorporating such regularization loss is helpful. The entire loss function used for 4DGS training is: $\mathcal{L}_\text{4DGS} = |\hat{I}_{M,t} - I_{M,t}| + \mathcal{L}_\text{TV}$.

As a result, we obtain the optimized 4D Gaussians $\left\{\mathcal{G}_{\text{canon}}^\text{opt},\mathcal{E}^\text{opt}(\mathcal{G}_{\text{canon}}, t), \mathcal{D}^\text{opt}\right\}$, which represent the editing target scene. The static component $\mathcal{G}_{\text{canon}}^\text{opt}$ serves as the main editing target in Sec.~\ref{subsec::4.2}--\ref{subsec::4.3}. In Sec.~\ref{subsec::4.2}, we edit the static component by modifying only the images corresponding to the first timestep, ensuring efficient editing. In Sec.~\ref{subsec::4.3}, we refine the edited static component $\mathcal{G}_{\text{canon}}^\text{edit}$ to better align with the original deformation field $\left\{\mathcal{E}^\text{opt}(\mathcal{G}_{\text{canon}}, t),\mathcal{D}^\text{opt}\right\}$ using score-based temporal refinement, mitigating potential motion artifacts. A more detailed training setup follows~\cite{ref_10_4dgs}.

\subsection{Stage 1: Efficient Dynamic Scene Editing with Static 3D Gaussians}
\label{subsec::4.2}
In Sec.~\ref{subsec::4.1}, we obtained the optimized canonical 3D Gaussians $\mathcal{G}_{\text{canon}}^\text{opt}$, which models the explicit appearance and geometry of a 3D scene, along with the optimized dynamic components $\mathcal{E}^\text{opt}(\mathcal{G}_{\text{canon}}, t)$ and $\mathcal{D}^\text{opt}$. For efficient dynamic scene editing, we perform editing only on $\mathcal{G}_{\text{canon}}^\text{opt}$, which is minimal but sufficient information for visual editing of the dynamic scene as shown in Fig.~\ref{fig:method}.

To generate supervision images for editing the $\mathcal{G}_{\text{canon}}^\text{opt}$, we extract a subset of multiview images fixed at the initial timestep and then edit them using InstructPix2Pix~\cite{ref_1_ip2p}. Subsequently, we edit optimized canonical 3D Gaussians $\mathcal{G}_{\text{canon}}^\text{opt}$ with an L1 RGB loss supervising the edited images. Compared to the latest 4D editing method Instruct 4D-to-4D~\cite{ref_9_i4d24d}, which requires editing $T\!\times\!\mathcal{M}$ images---where $T$ is the number of video timesteps and $\mathcal{M}$ is the number of cameras---through iterative dataset updates, our approach significantly reduces the computation required to address the editing of the dynamic scene. Moreover, this approach allows rapid transitions to the edited result, regardless of the number of timesteps $T$ of the dynamic scene. After completing the 3D Gaussian editing process, we obtain a \textbf{\textit{pseudo-edited dynamic scene}} $\left\{\mathcal{G}_{\text{canon}}^\text{edit},\mathcal{E}^\text{opt}(\mathcal{G}_{\text{canon}}, t), \mathcal{D}^\text{opt}\right\}$, which is obtained by simply recombining the edited canonical 3D Gaussians $\mathcal{G}_{\text{canon}}^\text{edit}$ with the original Gaussian deformation field $\left\{\mathcal{E}^\text{opt}(\mathcal{G}_{\text{canon}}, t), \mathcal{D}^\text{opt}\right\}$.

To ensure spatial consistency of the 3D Gaussian editing process, we utilize Coherent-IP2P~\cite{ref_9_i4d24d, ref_20_mvdream, ref_21_tuneavideo}, which replaces the 2D convolutional layer (self-attention module) with a 3D convolutional layer (cross-attention module), similar to Instruct 4D-to-4D (by reusing the original parameters of kernels). As shown in Fig.~\ref{fig:ablation}, this encourages collaborative editing among images within the multiview subset, preventing the results from becoming blurry. The entire editing process for the static canonical 3D Gaussian editing can be completed within a few tens of minutes by editing only multiview images of a single timestep and performing a few hundred 3DGS editing iterations.

\subsection{Stage 2: Refinement using Score Distillation for Temporal Alignment}
\label{subsec::4.3}
\begin{figure}[!t]
\centering
    \includegraphics[width=1.0\columnwidth]{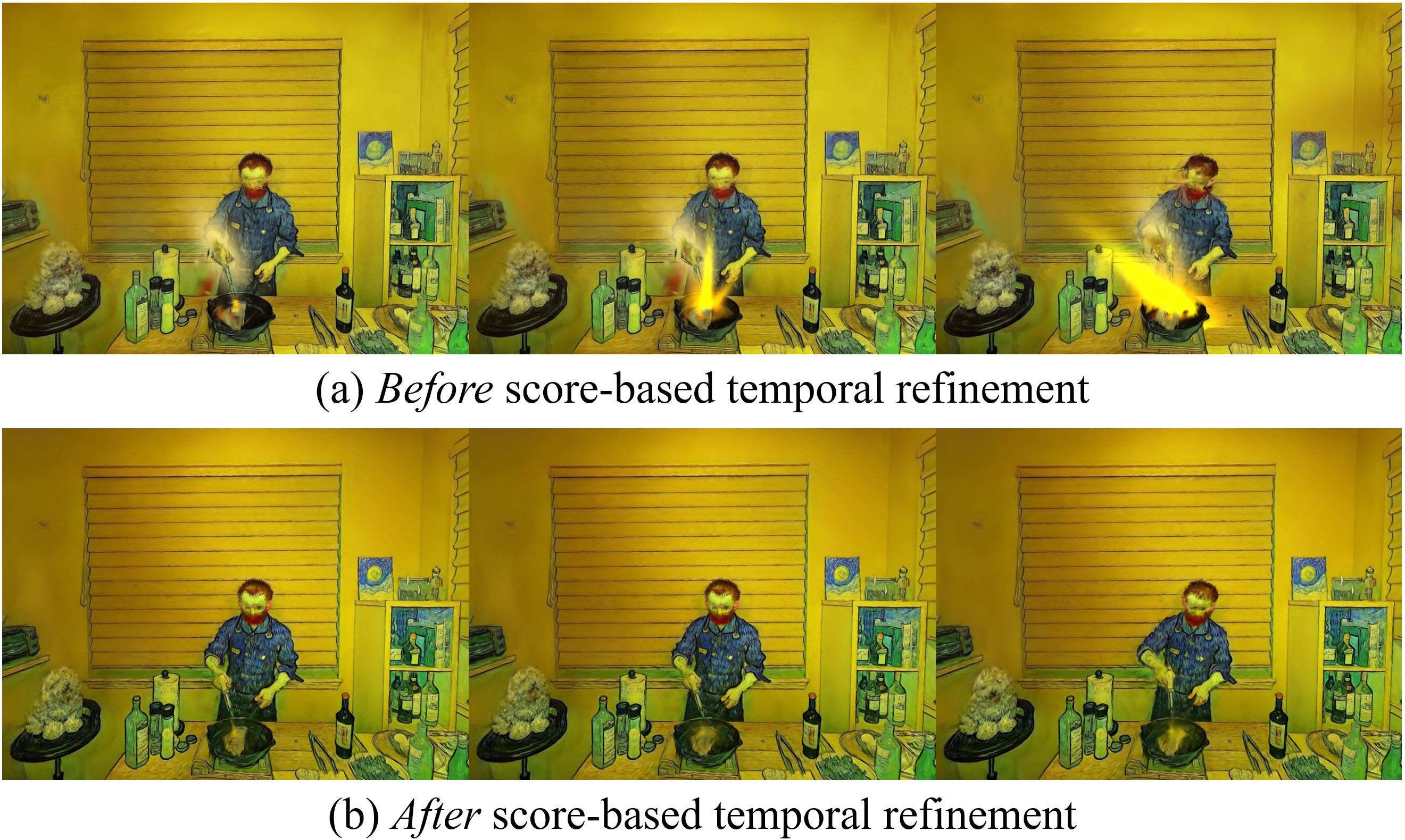}
    \vspace{-7mm}
    \caption{\textbf{Effectiveness of score-based temporal refinement}: Score-based temporal refinement effectively resolves misalignment between the canonical 3D Gaussians and the original deformation field that arises during the 3D Gaussian editing process. Without requiring additional 2D image updates, this process completes dynamic scene editing within a few hundred iterations.}
    \label{fig:scoredistillation}
\vspace{-3mm}
\end{figure}

After the first editing stage proposed in Sec.~\ref{subsec::4.2}, the \textit{pseudo-edited dynamic scene} $\left\{\mathcal{G}_{\text{canon}}^\text{edit},\mathcal{E}^\text{opt}(\mathcal{G}_{\text{canon}}, t), \mathcal{D}^\text{opt}\right\}$ exhibits severe motion artifacts, as shown in Fig.~\ref{fig:scoredistillation} \red{(a)}. The primary cause is the slight shift in the positions $p$ of Gaussian primitives in $\mathcal{G}_{\text{canon}}^\text{opt}$ during the 3D Gaussian editing process, which results in discrepancies between the queried embedding voxel feature $f_h$ and those of the original dynamic scene. Moreover, only the Spherical Harmonics (SH) colors on the surface visible at the initial timestep are updated. As a result, if the Gaussian primitives in \textit{pseudo-edited} $\mathcal{G}_{\text{canon}}^\text{edit}$ rotate at later timesteps, unedited SH values that were previously hidden may become exposed, leading to artifacts. Therefore, we introduce a temporal refinement stage to resolve the misalignment between the original deformation field $\left\{\mathcal{E}^\text{opt}(\mathcal{G}_{\text{canon}}, t), \mathcal{D}^\text{opt}\right\}$ and the edited canonical 3D Gaussians $\mathcal{G}_{\text{canon}}^\text{edit}$.

To perform the refinement stage efficiently without editing additional dataset images, we employ the score distillation mechanism ~\cite{ref_18_dreamfusion}. Since the dynamic scene is edited using multiple 2D images generated by IP2P, the prior of the 2D diffusion model (\ie, IP2P) can be distilled into the 4D dynamic scene. The editing process can be continued using the noise prediction loss (\ie score) obtained from each IP2P inference as Eqs.~\ref{eq::eq1} and~\ref{eq::eq2}. Since we just use score distillation for editing refinement rather than generation or editing from scratch, this stage can be completed with a smaller number of iterations. Consequently, our approach is relatively less affected by inherent issues of Score Distillation Sampling (SDS), such as \emph{Janus problem}~\cite{ref_20_mvdream}.

Similar to Sec.~\ref{subsec::4.2}, we apply Coherent-IP2P with the diffusion prior $\theta$ and observe that it reduces blurring effects and enhances qualitative performance compared to the original IP2P. At each refinement iteration, we rendered $B$ images of the \textit{pseudo-edited dynamic scene} $\tilde{I} = \left\{ \hat{I}_i = S(M_i, \mathcal{G}_{\text{def},t_i}^\text{edit}) \right\}_{i=1}^B$ using random camera matrices $\left\{ M_i \right\}_{i=1}^B$ and random timesteps $\left\{ t_i \right\}_{i=1}^B$ as input for Coherent-IP2P, where $S$ denotes the rendering process of the 3DGS (subscripts $M$ and $t$ on $\hat{I}$ omitted for simplicity). We optimize $\mathcal{G}_{\text{canon}}^\text{edit}$ using the following SDS loss to obtain the refined 3D Gaussians $\mathcal{G}_{\text{canon}}^\text{ref}$:
\vspace{-2mm}
\begin{equation}
    \scriptsize
    \nabla_{\mathcal{G}_{\text{canon}}^{\text{edit}}} \mathcal{L}_{\text{SDS}} = \mathbb{E}_{t, \tilde{t}, \epsilon, \mathcal{M}} \left[ \left( \epsilon_{\theta} \left( \tilde{I}, c_I, c_T; t, \tilde{t}, \mathcal{M} \right) - \epsilon \right) \frac{\partial \tilde{I}}{\partial \mathcal{G}_{\text{canon}}^{\text{edit}}} \right],
\label{eq::eq1}
\end{equation}

\vspace{-4mm}
\begin{equation}
    \footnotesize
    \begin{split}
        \epsilon_{\theta}(\tilde{I}, c_I, c_T) = \epsilon_{\theta}(\tilde{I}, \emptyset, \emptyset) + s_I \left( \epsilon_{\theta}(\tilde{I}, c_I, \emptyset) - \epsilon_{\theta}(\tilde{I}, \emptyset, \emptyset) \right) \quad \\ + s_T \left( \epsilon_{\theta}(\tilde{I}, c_I, c_T) - \epsilon_{\theta}(\tilde{I}, c_I, \emptyset) \right)
    \end{split}
\label{eq::eq2}
\end{equation}
, where $\tilde{t}$ is diffusion timestep, $\epsilon$ is diffusion noise, $c_I = \left\{ I_i \right\}_{i=1}^B$ is original dataset images, $c_T$ is user instruction, $s_I$ and $s_T$ are Classifier-Free-Guidance~\cite{ref_43_classifierfree} scale for $c_I$ and $c_T$, $\epsilon_{\theta}(\tilde{I}, c_I, c_T)$ is Coherent-IP2P denoiser networks including VAE~\cite{ref_44_vae}. This score-based guidance encourages a set of rendered 2D images from the \textit{pseudo-edited} 4D Gaussians at arbitrary timesteps $\tilde{I}$ to resemble the edited images that IP2P would generate based on the $c_I$ and $c_T$, thereby effectively refining motion artifacts. As a result, we obtain refined canonical 3D Gaussians $\mathcal{G}_{\text{canon}}^\text{ref}$ that aligns well with the original deformation field $\left\{\mathcal{E}^\text{opt}(\mathcal{G}_{\text{canon}}, t), \mathcal{D}^\text{opt}\right\}$ while maintaining the edited appearance. After the refinement stage, we obtain a completely edited 4D dynamic scene which is represented as $\left\{\mathcal{G}_{\text{canon}}^\text{ref},\mathcal{E}^\text{opt}(\mathcal{G}_{\text{canon}}, t), \mathcal{D}^\text{opt}\right\}$.
\section{Experiments}
\subsection{Experimental Setup}
\paragraph{Datasets.}
We use DyNeRF~\cite{ref_37_neural3dvideo} and Technicolor~\cite{ref_71_technicolor}, a real-world multiview video dataset, to train and edit 4D dynamic scenes. The DyNeRF dataset includes six 10-second video sequences captured at 30 fps by 15 to 20 cameras with a face-forward perspective. Technicolor includes a wider variety of motion and scenarios, captured with 16 cameras. For comparison with the baseline~\cite{ref_1_ip2p}, we trim the videos into 50-frame-long segments. We have also included the results on monocular datasets~\cite{ref_73_hypernerf,ref_72_dycheck} in the supplementary.

\vspace{-4mm}\paragraph{Baselines.}
We conduct a qualitative and quantitative comparison with Instruct 4D-to-4D~\cite{ref_9_i4d24d}, the only prior work addressing instruction-guided 4D dynamic scene editing. Instruct 4D-to-4D utilizes NeRFPlayer~\cite{ref_22_nerfplayer} as its backbone 4D representation and employs an iterative dataset update method, which involves editing all 2D images used for synthesizing the dynamic scene. It utilizes optical flow-based warping~\cite{ref_53_raft} and depth-based warping to ensure consistency across all edited 2D images. To alleviate the time-consuming dataset update process, Instruct 4D-to-4D employs two GPUs in parallel: one for the dataset update thread and the other for the dynamic scene editing thread.
\vspace{-5mm}\paragraph{Implementation Details.}
In Sec.~\ref{subsec::4.1}, we follow the experimental settings of~\cite{ref_10_4dgs}. Throughout the experiments utilizing InstructPix2Pix~\cite{ref_1_ip2p}, we set the CFG~\cite{ref_43_classifierfree} scales for image condition and text instruction to 1.2 and 8.5 to 10.5, respectively. In the 3D Gaussian editing stage (Sec.~\ref{subsec::4.2}), we train for 800 to 1000 iterations, depending on the editing style. For the score-based refinement stage (Sec.~\ref{subsec::4.3}), an average of 800 iterations is sufficient to complete the dynamic scene editing successfully. All experiments are conducted using a single NVIDIA A40 GPU.

\vspace{-1mm}
\subsection{Results}
\vspace{-1mm}
\paragraph{Quantitative Results.} 
To quantitatively evaluate the visual quality of the edited dynamic scene, we measure PSNR, SSIM~\cite{ref_38_ssim}, and LPIPS~\cite{ref_39_lpips} between the 2D multiview images used as supervision for dynamic scene editing and the images rendered from the edited dynamic scene using the corresponding camera parameters. Additionally, to assess how well the edited dynamic scene aligns with the input instruction, we also measure CLIP~\cite{ref_40_clip} similarity.

Table~\ref{tab:comparison_metrics} presents a quantitative comparison of our method, Instruct-4DGS, and the baseline, Instruct 4D-to-4D, on DyNeRF. While our method shows slightly worse PSNR and SSIM in some cases, this is expected due to our efficient editing strategy. Unlike the baseline, which directly optimizes pixel-level accuracy using all edited images as training targets, our approach optimizes the dynamic scene using only instructions, without additional image editing during the temporal refinement stage. Consequently, pixel-wise accuracy (\ie, PSNR and SSIM) may be worse, but our method demonstrates superior perceptual quality, as shown in the consistently lower LPIPS across all cases. Additionally, our approach excels in instruction-following fidelity, achieving higher CLIP similarity than the baseline. Notably, our method accomplishes this 2--3 times faster while requiring fewer GPUs, making it significantly more efficient for real-world applications.

Table~\ref{tab:computing_time_comparison} compares the efficiency of the baseline and our method. In terms of editing time, our method completes editing 2--3 times faster while using only a single GPU, whereas the baseline requires two identical GPUs. This speed advantage could potentially become more pronounced as the number of timesteps in the dynamic scene increases. These results demonstrate that our method achieves efficiency by leveraging the static-dynamic separability of 4DGS and employing score-based temporal refinement, enabling significantly faster dynamic scene editing without extensive, time-consuming dataset updates.

\begin{table}[!t]
\centering
\footnotesize
\resizebox{1.0\columnwidth}{!}{
\begin{tabular}{c|c|c|c|c|c}
\toprule
Instruction & Method & PSNR$\uparrow$ & SSIM$\uparrow$ & $\text{LPIPS}_{\scriptscriptstyle \text{VGG}}$$\downarrow$ & CLIP sim.$\uparrow$ \\
\hline
\multirow{2}{*}{\textit{Statue}}          & I4D24D  & 18.73 & 0.713 & 0.567 & 0.202 \\ \cline{2-6} 
                                          & Ours    & \textbf{21.41} & \textbf{0.829} & \textbf{0.259} & \textbf{0.220} \\ 
\hline
\multirow{2}{*}{\makecell{\textit{Roman}\\\textit{Sculpture}}} & I4D24D  & \textbf{24.24} & \textbf{0.865} & 0.372 & 0.229 \\ \cline{2-6}
                                          & Ours    & 18.69 & 0.801 & \textbf{0.329} & \textbf{0.252} \\ 
\hline
\multirow{2}{*}{\makecell{\textit{Wood}\\\textit{Sculpture}}}  & I4D24D  & \textbf{18.23} & 0.631 & 0.535 & 0.258 \\ \cline{2-6} 
                                          & Ours    & 17.64 & \textbf{0.718} & \textbf{0.321} & \textbf{0.276} \\ 
\hline
\multirow{2}{*}{\textit{(Average)}}       & I4D24D  & \textbf{20.40} & 0.736 & 0.491 & 0.230 \\ \cline{2-6} 
                                          & Ours    & 19.25 & \textbf{0.783} & \textbf{0.303} & \textbf{0.249} \\ 
\bottomrule
\end{tabular}
}
\vspace{-2mm}
\caption{\textbf{Quantitative comparison of editing quality}: Comparison of performance metrics between Instruct 4D-to-4D (I4D24D) and our Instruct-4DGS (Ours) under various editing instructions on DyNeRF. Higher values indicate better performance for PSNR, SSIM, and CLIP similarity; lower values are better for $\text{LPIPS}_{\text{VGG}}$.}
\label{tab:comparison_metrics}
\end{table}

\begin{table}[!t]
\centering
\resizebox{1.0\columnwidth}{!}{
    \begin{tabular}{c|c|c}
    \toprule
    Method & Computing units & Avg. editing time \\
    \hline
    Instruct 4D-to-4D~\cite{ref_9_i4d24d} & 2 GPUs & 2 hours \\
    \hline
    Instruct-4DGS & \textbf{1 GPU} & \textbf{40 minutes} \\
    \bottomrule
    \end{tabular}
    }
\vspace{-2mm}
\caption{\textbf{Quantitative comparison of editing efficiency}: Our proposed Instruct-4DGS significantly reduces editing time even with fewer GPU resources compared to the baseline.}
\vspace{-3mm}
\label{tab:computing_time_comparison}
\end{table}

\begin{figure}[!t]
\centering
    \includegraphics[width=\linewidth]{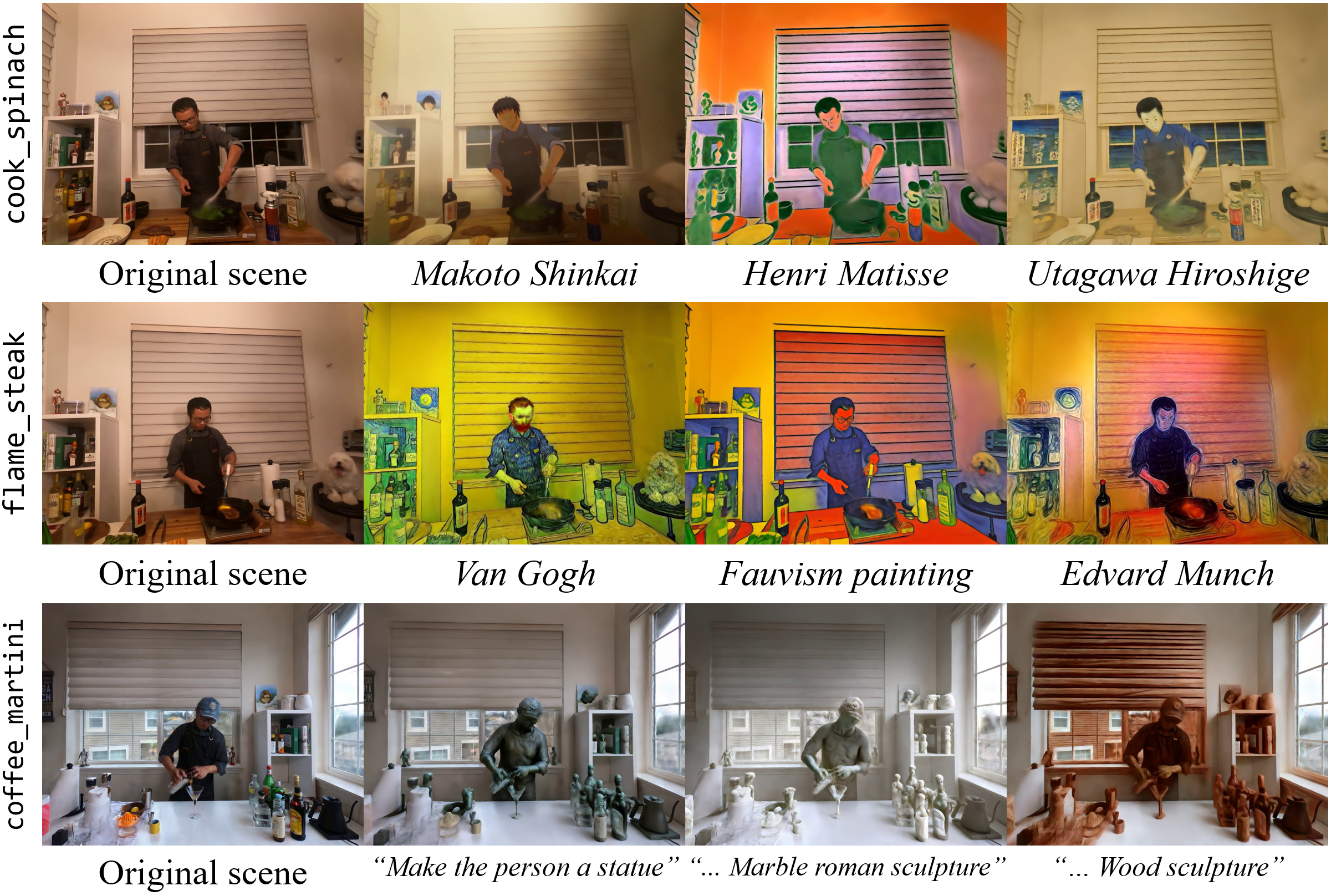}
    \vspace{-6mm}
    \caption{\textbf{Qualitative results across various editing styles}: Editing results of the scenes \emph{cook\_spinach}, \emph{flame\_steak}, and \emph{coffee\_martini} scenes from DyNeRF. Instruct-4DGS successfully edits dynamic scenes closely following the given user instructions.}
    \vspace{-3mm}
    \label{fig:qual_1}
\end{figure}

\vspace{-5mm}\paragraph{Qualitative Results.} 
Our qualitative results are shown in Fig.~\ref{fig:qual_1}. Our dynamic scene editing method effectively follows various editing styles based on the provided instructions. Leveraging the capabilities of 4DGS~\cite{ref_10_4dgs}, each rendered image exhibits high fidelity, accurately capturing the target details. Moreover, the rendered video output of our edited dynamic scenes maintains smooth motion. 

\begin{figure*}[h]
\centering
    \includegraphics[width=1.0\linewidth]{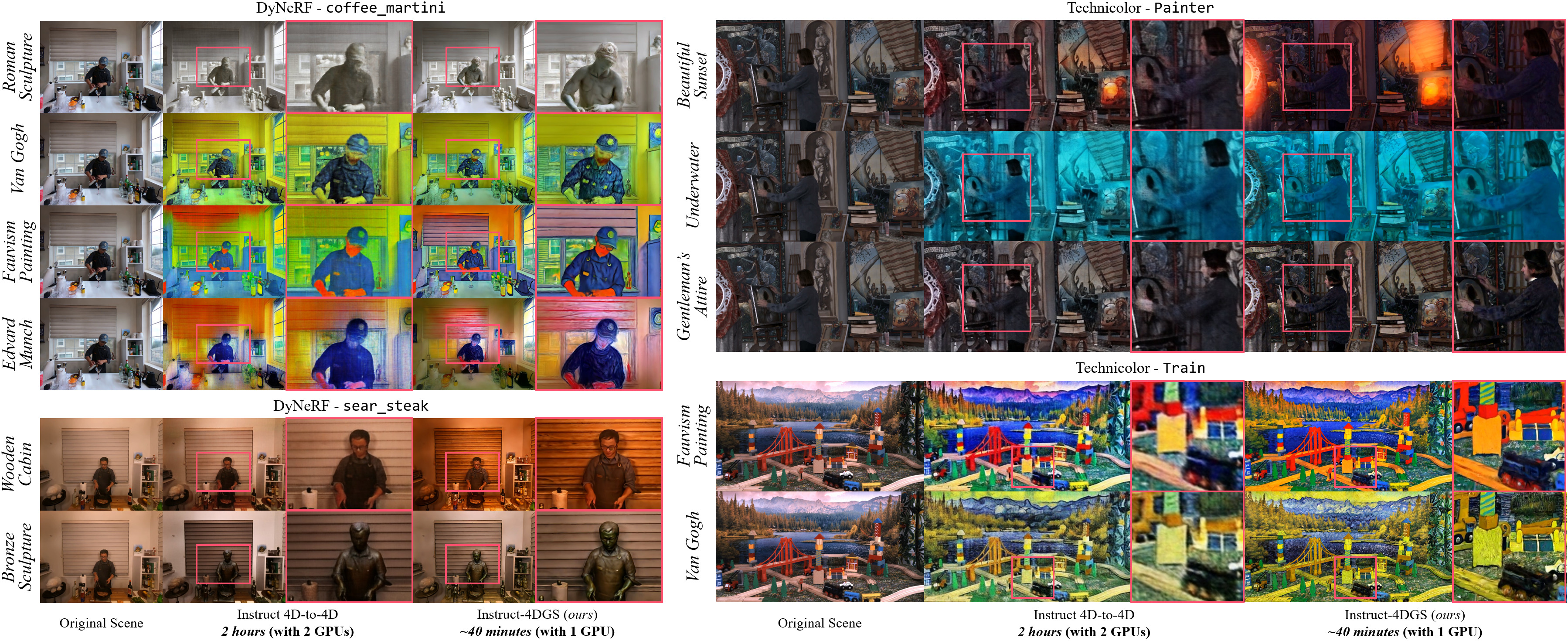}
    \vspace{-7mm}
    \caption{\textbf{Qualitative comparison of visual quality}: We compare our method with the baseline~\cite{ref_9_i4d24d} on DyNeRF~\cite{ref_37_neural3dvideo} \emph{coffee\_martini} and \emph{sear\_steak} scenes, as well as Technicolor~\cite{ref_71_technicolor}'s \emph{Painter} and \emph{Train} scenes. See supplementary for more results.}
    \label{fig:qual_2}
\vspace{-4mm}
\end{figure*}

\begin{figure}[!t]
\centering
    \includegraphics[width=\columnwidth]{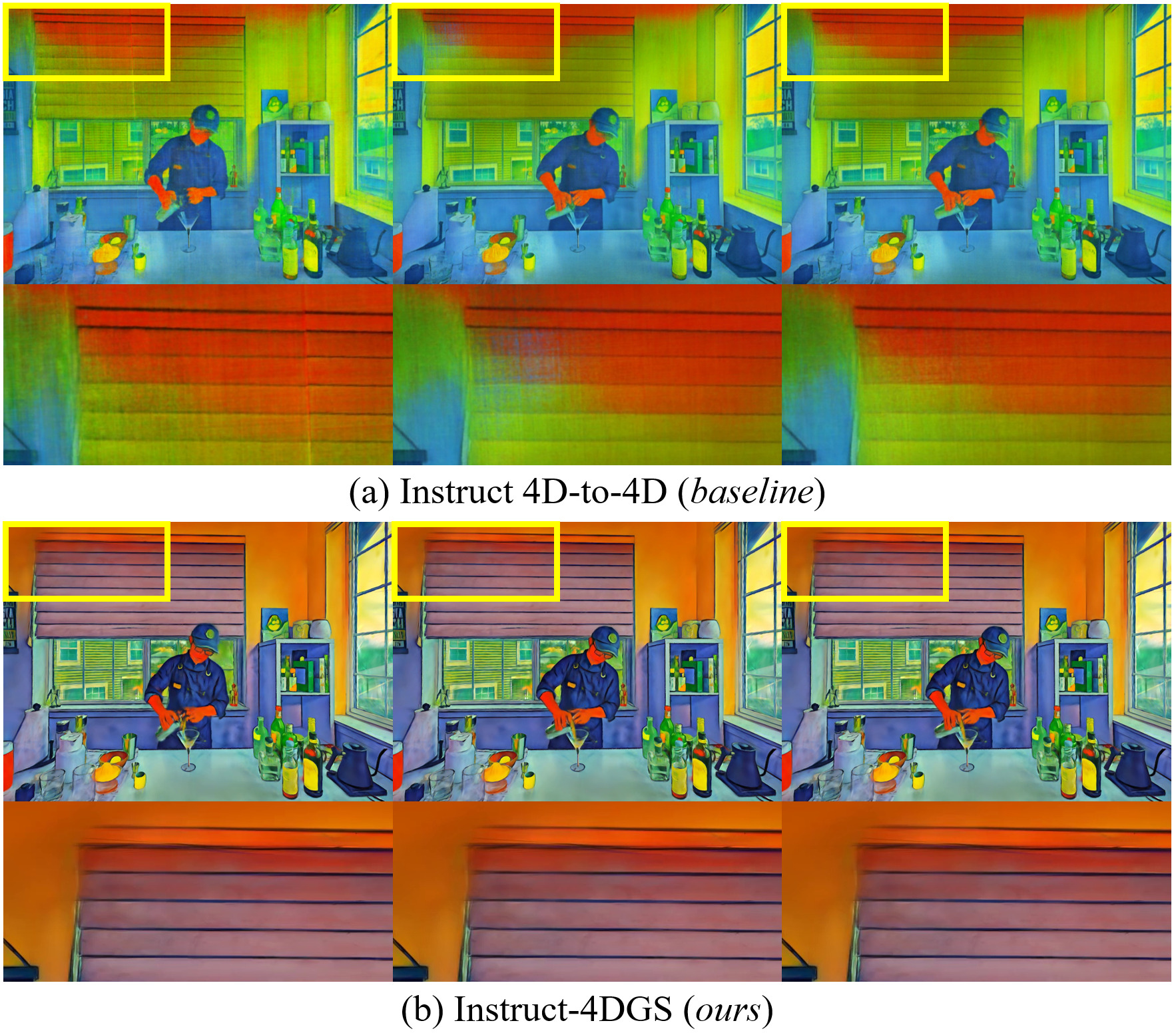}
    \vspace{-7mm}
    \caption{\textbf{Qualitative comparison of temporal consistency}: The baseline shows noticeable flickering artifacts across timesteps. In contrast, Instruct-4DGS effectively avoids such artifacts by editing only the static component with score-based temporal refinement.}
    \label{fig:qual_3}
\vspace{-4mm}
\end{figure}

Qualitative comparison with the baseline is in Fig.~\ref{fig:qual_2} and Fig.~\ref{fig:qual_3}. As shown in the zoomed-in images of Fig.~\ref{fig:qual_2}, our method produces a high-quality edited dynamic scene with less noise and blurry artifacts compared to the baseline. Furthermore, as shown in Fig.~\ref{fig:qual_3}, a comparison of images across multiple timesteps from a fixed camera reveals that the baseline exhibits a noticeable flickering effect. These results indicate that, although the baseline attempts to ensure consistency across all 2D image edits, it falls short of achieving full temporal consistency. In comparison, our method avoids such artifacts by editing the dynamic scene across the temporal dimension through score refinement. It is worth emphasizing that these higher-quality results are achieved 2–3 times faster than the baseline.

\begin{figure}[!t]
\centering
    \includegraphics[width=\columnwidth]{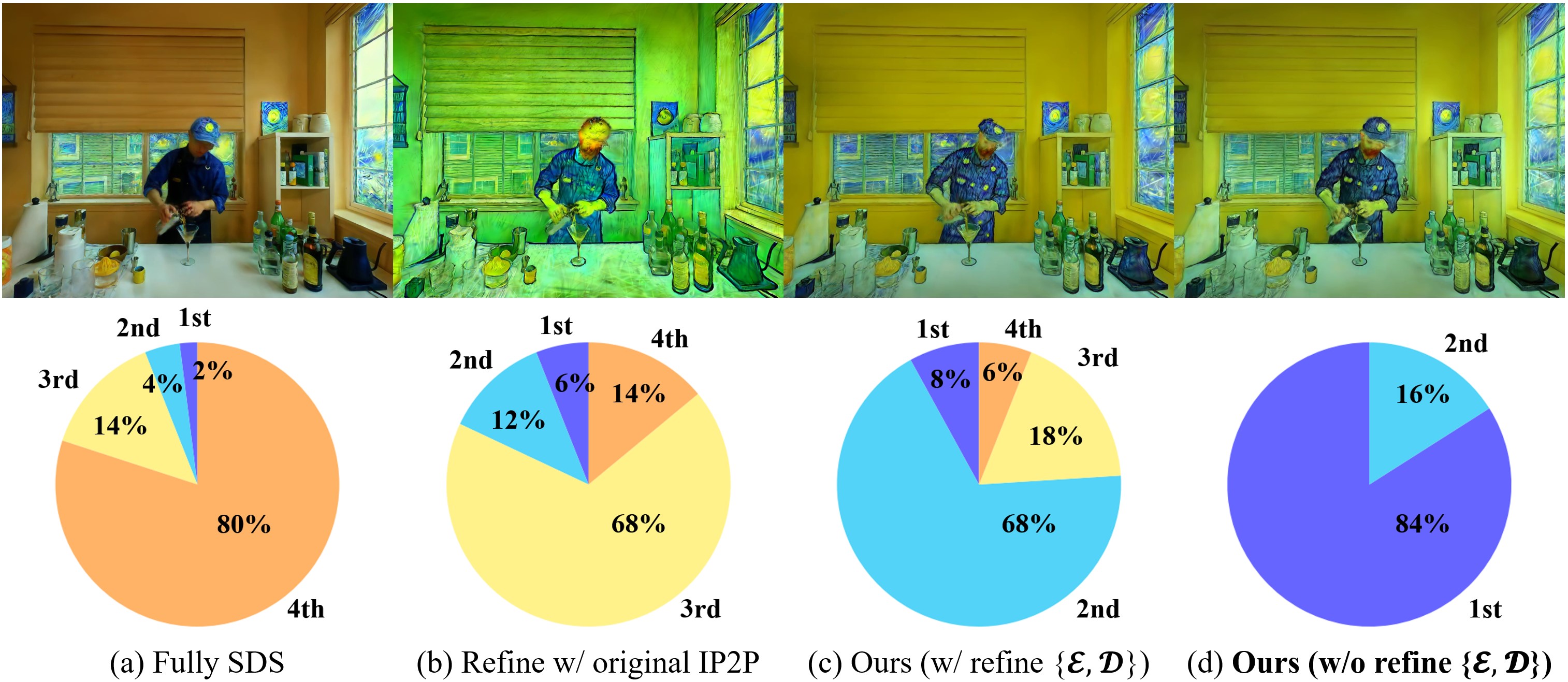}
    \vspace{-6mm}
    \caption{\textbf{Ablation study of the dynamic scene editing method}: Each pie chart shows the proportion of user preferences (1st-4th ranks) for each method variant. Our proposed method (denoted as ``Ours (w/o refine $\{\mathcal{E}, \mathcal{D}\}$)'') achieves the highest preference score.
    }
    \label{fig:ablation}
\end{figure}

\vspace{-5mm}\paragraph{Ablation Studies.} 
We conduct an ablation study to evaluate the impact of each design choice in our method, particularly their contributions to efficiency and quality. The qualitative and user study results are shown in Fig.~\ref{fig:ablation}. We recruited 50 participants of varying demographics, collecting a total of 50 preference rankings on the editing quality of videos generated by the four method variants.

First, we examine dynamic scene editing using only score-based editing, without 3D Gaussian editing (denoted as ``Fully SDS''). As shown in Fig.~\ref{fig:ablation} \red{(a)}, this approach preserves smooth motion but fails to ensure sufficient instruction alignment, leading to low-fidelity results. In contrast, incorporating the 3D Gaussian editing stage significantly improves fidelity while enabling effective motion refinement in the temporal refinement process. This highlights the importance of direct supervision via edited 2D images in maintaining fidelity and quality.

To mitigate inherent issues of SDS such as \emph{Janus problem}~\cite{ref_20_mvdream}, and to provide stable guidance for the score-based temporal refinement, we employ Coherent-IP2P. To validate this choice, we compare results by refining the \textit{pseudo-edited dynamic scene} with the original IP2P (denoted as ``Refine w/ original IP2P''). As shown in Fig.~\ref{fig:ablation} \red{(b)}, using the original IP2P for temporal refinement leads to severe visual artifacts and low-quality outputs, whereas Coherent-IP2P preserves details and retains the scene’s semantics. This confirms that Coherent-IP2P mitigates noisy guidance and blurry artifacts by enabling information sharing among images within the same batch.

Lastly, to evaluate the effectiveness of refining the deformation field, we compare our final method (denoted as ``Ours (w/o refine $\{\mathcal{E}, \mathcal{D}\}$)'')---which refines only the static 3D Gaussians, excluding the deformation field---with ``Ours (w/ refine $\{\mathcal{E}, \mathcal{D}\}$)''. As shown in Fig.~\ref{fig:ablation} \red{(c)}, refining the deformation field introduces temporal inconsistencies and motion artifacts. In contrast, our final method effectively preserves temporal coherence while maintaining high editing fidelity. These results indicate that refining the deformation field does not contribute positively to dynamic scene editing and can instead introduce undesirable distortions.
\section{Conclusion and Limitations}
\vspace{-2mm}
\paragraph{Conclusion.} We proposed Instruct-4DGS, an efficient 4D dynamic scene editing framework leveraging 4D Gaussian Splatting (4DGS) and a score distillation mechanism. Exploiting the static-dynamic separability of 4DGS, our approach edits only static canonical components and refines motion artifacts, significantly enhancing editing speed and efficiency. Score distillation effectively transfers generative priors into 4D space, offering an efficient alternative to the conventional RGB loss, which requires updating additional 2D images. Experimental results demonstrated superior visual quality and editing efficiency compared to the baseline.

\vspace{-4mm}
\paragraph{Limitations.} Our method relies on IP2P's capabilities, cannot directly edit motion, requires segmentation for partial edits, and may show motion artifacts due to limitations of the 4D representation, even after temporal refinement.

\vspace{-4mm}
\paragraph{Acknowledgements.} This work was supported by Institute of Information \& Communications Technology Planning \& Evaluation (IITP) grant funded by the Korea government (MSIT) (RS-2024-00439020, Developing Sustainable, Real-Time Generative AI for Multimodal Interaction, SW Starlab).

{
    \small
    \bibliographystyle{ieeenat_fullname}
    \bibliography{main}
}

\clearpage
\setcounter{page}{1}
\maketitlesupplementary

\section{Additional Qualitative Results}

\subsection{Results on Monocular Datasets}
\begin{figure*}[!t]
\centering
    \includegraphics[width=0.85\linewidth]{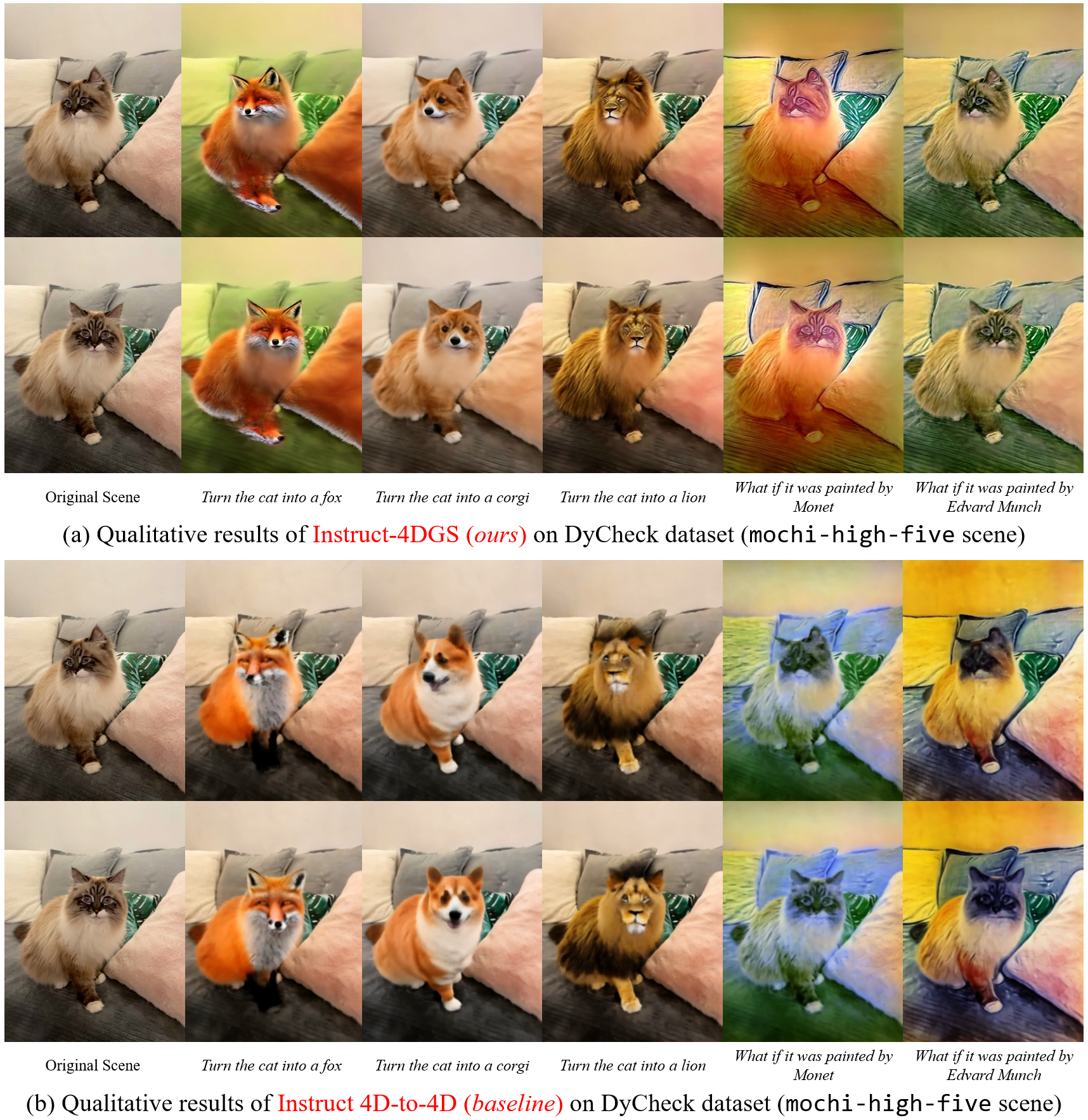}
    \caption{\textbf{Qualitative comparison of visual quality on the DyCheck~\cite{ref_72_dycheck} dataset (a \emph{monocular} dataset)}: We compare our method, Instruct-4DGS (\emph{ours}), with the baseline, Instruct 4D-to-4D~\cite{ref_9_i4d24d} (\emph{baseline}), on the \emph{mochi-high-five} scene from the DyCheck dataset.}
    \label{fig:suppl_dycheck}
\end{figure*}

\begin{figure*}[!t]
\centering
    \includegraphics[width=0.85\linewidth]{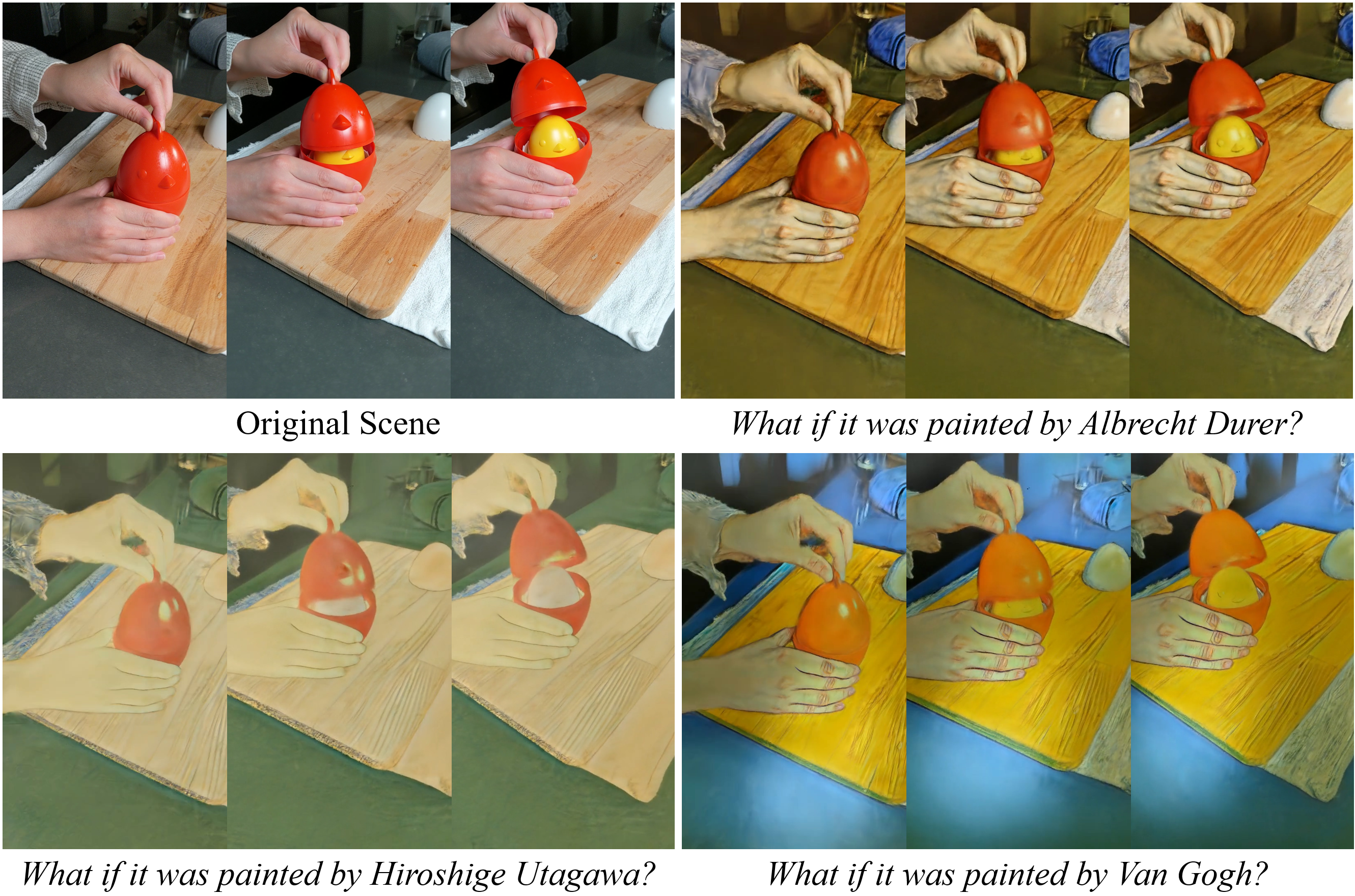}
    \caption{\textbf{Qualitative results of our Instruct-4DGS on the HyperNeRF~\cite{ref_73_hypernerf} dataset (a \emph{monocular} dataset)}: We evaluate our method on the \emph{Interp\_chickchicken} scene from the HyperNeRF dataset.}
    \label{fig:suppl_hypernerf}
\end{figure*}

While 4D dynamic scene editing typically relies on multiview video datasets to sufficiently capture spatio-temporal information, we evaluate our method on the DyCheck~\cite{ref_72_dycheck} and HyperNeRF~\cite{ref_73_hypernerf} datasets to explore its potential applicability to monocular video inputs. For these monocular datasets, we cannot obtain edited multiview supervision images for editing canonical 3D Gaussians. Therefore, we skip Stage 1 (described in Sec.~\ref{subsec::4.2}) and only apply Stage 2, the score-based temporal refinement (described in Sec.~\ref{subsec::4.3}). Figure~\ref{fig:suppl_dycheck} presents a comparison between Instruct 4D-to-4D (\emph{baseline}) and Instruct-4DGS (\emph{ours}) on the DyCheck dataset, while Fig.~\ref{fig:suppl_hypernerf} shows qualitative results of our method on the HyperNeRF dataset. Our Instruct-4DGS produces plausible dynamic scene editing results even on monocular datasets, and we expect the performance to further improve as techniques for reconstructing 4D Gaussians from monocular videos and editing with the SDS mechanism continue to advance.

\subsection{Results with Varying Camera Poses}
\begin{figure*}[t]
\centering
    \includegraphics[width=1.0\linewidth]{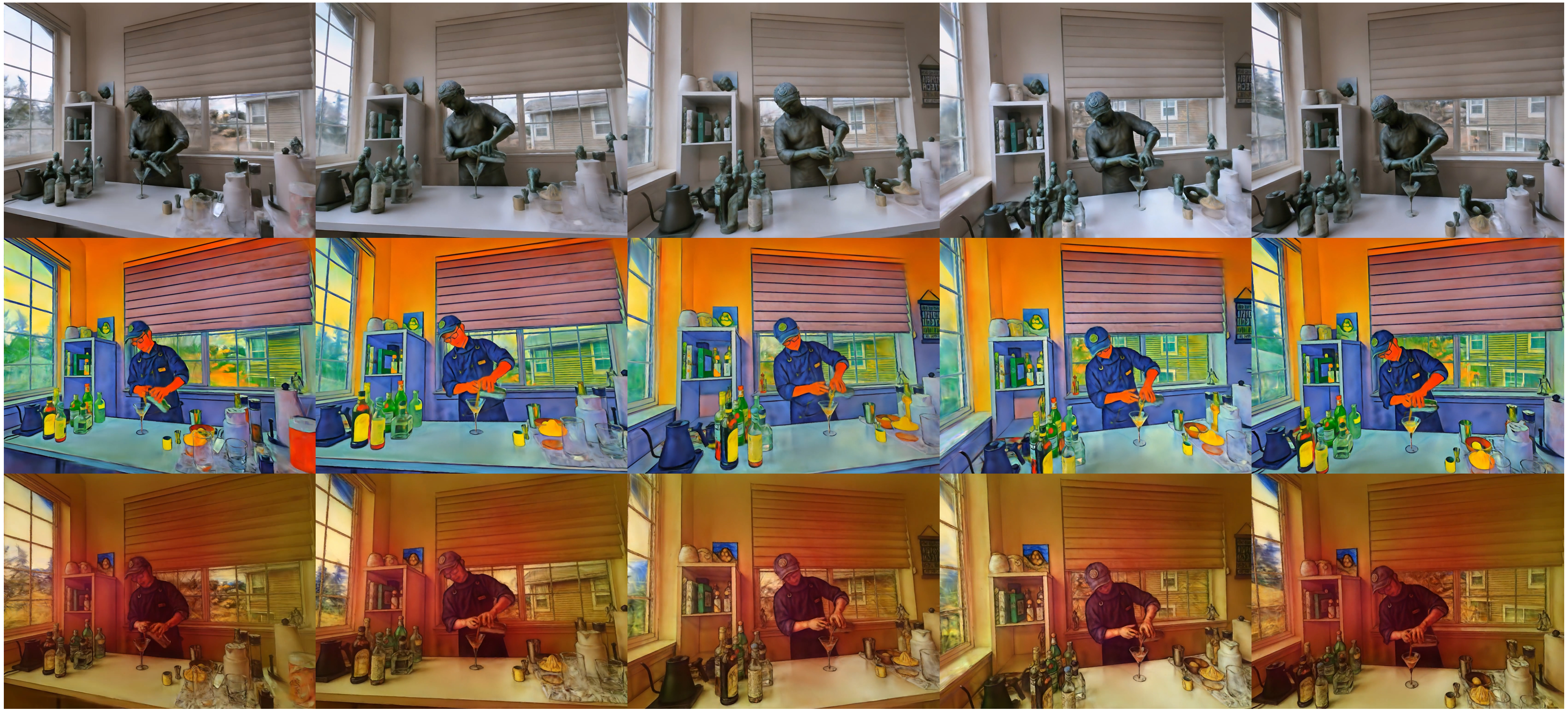}
    \caption{\textbf{Qualitative results of our Instruct-4DGS under various camera poses on the DyNeRF~\cite{ref_37_neural3dvideo} dataset}: We render the edited dynamic scene from novel camera poses to evaluate the spatial consistency of our method. Our Instruct-4DGS produces view-consistent and geometrically plausible results.}
    \label{fig:suppl_dynerf}
\end{figure*}

To further assess the spatial consistency and quality of our edited 4D Gaussian representations, we render the edited dynamic scenes from the DyNeRF~\cite{ref_37_neural3dvideo} dataset under various camera poses. As shown in Fig.~\ref{fig:suppl_dynerf}, the results produced by our Instruct-4DGS maintain plausible geometry and appearance across different viewpoints.

\section{Full Set of Editing Instructions}
Here, we provide the full set of editing instructions used for our dynamic scene editing experiments.

We used \emph{``Make the person a statue''}, \emph{``Make the person a marble Roman sculpture''}, and \emph{``Make the person a wood sculpture''} for Tab.~\ref{tab:comparison_metrics}.

We used \emph{``What if it was painted by \{Makoto Shinkai, Henri Matisse, Utagawa Hiroshige, Van Gogh, Edvard Munch\}?''}, \emph{``Make it a Fauvism painting''}, \emph{``Make the person a statue''}, \emph{``Make the person a marble Roman sculpture''}, and \emph{``Make the person a wood sculpture''} for Fig.~\ref{fig:qual_1}.

We used \emph{``Make the person a marble Roman sculpture''}, \emph{``What if it was painted by \{Van Gogh, Edvard Munch\}?''}, \emph{``Make it a Fauvism painting''}, \emph{``Make this a cozy wooden cabin bar with soft lighting and rustic decorations''}, \emph{``Turn the man into a bronze sculpture''}, \emph{``Add a beautiful sunset''}, \emph{``Make it underwater''}, \emph{``Give him a Victorian gentleman’s attire''}, \emph{``Make it a Fauvism painting''}, and \emph{``What if it was painted by Van Gogh?''} for Fig.~\ref{fig:qual_2}

\end{document}